\documentclass[letterpaper]{article} %
\usepackage[]{aaai2026}  %
\usepackage{times}  %
\usepackage{helvet}  %
\usepackage{courier}  %
\usepackage[hyphens]{url}  %
\usepackage{graphicx} %
\usepackage{natbib}  %
\usepackage{caption} %
\usepackage{algorithm}
\usepackage{algorithmic}

\usepackage{newfloat}
\usepackage{listings}
\DeclareCaptionStyle{ruled}{labelfont=normalfont,labelsep=colon,strut=off} %
\floatstyle{ruled}
\newfloat{listing}{tb}{lst}{}
\floatname{listing}{Listing}
\usepackage{multirow}
\usepackage{amsmath}
\usepackage{amssymb}

\title{Toward Rich Video Human-Motion2D Generation}
\author{
    Ruihao Xi\equalcontrib\textsuperscript{\rm 1},
    Xuekuan Wang\equalcontrib\textsuperscript{\rm 1},
    Yongcheng Li\textsuperscript{\rm 1},
    Shuhua Li\textsuperscript{\rm 1}, \\
    Zichen Wang\textsuperscript{\rm 2},
    Yiwei Wang\textsuperscript{\rm 1},
    Feng Wei\textsuperscript{\rm 3},
    Cairong Zhao\textsuperscript{\rm 1}\thanks{Corresponding author.}
}
\affiliations{
    \textsuperscript{\rm 1}Department of Computer Science and Technology, Tongji University, China\\
    \textsuperscript{\rm 2}School of Mathematical Sciences, Tongji University, China\\
    \textsuperscript{\rm 3}College of Information Engineering, Shanghai Maritime University, China\\
    2254328@tongji.edu.cn, wxktongji@163.com, liyongcheng@tongji.edu.cn, 2152841@tongji.edu.cn, 2351343@tongji.edu.cn, JadeW@tongji.edu.cn, 202430310102@stu.shmtu.edu.cn, zhaocairong@tongji.edu.cn

}

\usepackage{bibentry}

\begin{document}

\maketitle

\begin{abstract}
Generating realistic and controllable human motions, particularly those involving rich multi-character interactions, remains a significant challenge due to data scarcity and the complexities of modeling inter-personal dynamics. To address these limitations, we first introduce a new large-scale rich video human motion 2D dataset (Motion2D-Video-150K) comprising 150,000 video sequences. Motion2D-Video-150K features a balanced distribution of diverse single-character and, crucially, double-character interactive actions, each paired with detailed textual descriptions. Building upon this dataset, we propose a novel diffusion-based rich video human motion2D generation (RVHM2D) model. RVHM2D incorporates an enhanced textual conditioning mechanism utilizing either dual text encoders (CLIP-L/B) or T5-XXL with both global and local features. We devise a two-stage training strategy: the model is first trained with a standard diffusion objective, and then fine-tuned using reinforcement learning with an FID-based reward to further enhance motion realism and text alignment. Extensive experiments demonstrate that RVHM2D achieves leading performance on the Motion2D-Video-150K benchmark in generating both single and interactive double-character scenarios.
\end{abstract}

\begin{links}
    \link{code}{https://github.com/FooAuto/Toward-Rich-Video-Human-Motion2D-Generation}
\end{links}

\section{Introduction}
\label{sec:intro}
\begin{figure*}
\begin{center}
\includegraphics[width=0.99\linewidth]{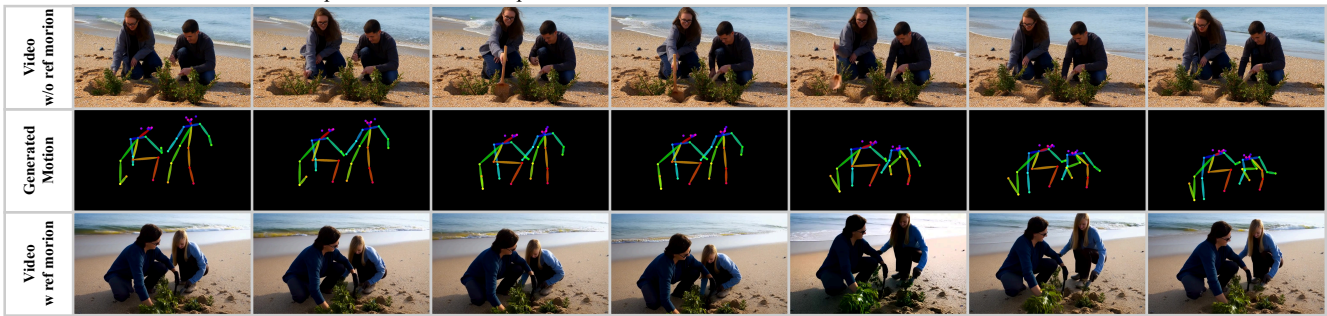} 
\end{center}
   \caption{Qualitative results illustrating our proposed text-to-motion generation (RVHM2D) and subsequent skeleton-driven video synthesis. We choose Wan2.1-T2V-14B to generate the videos. 
   These examples demonstrate RVHM2D's capability to produce textually-aligned and coherent motions that effectively drive high-quality video generation, showcasing significant improvements in motion fidelity and textual consistency over the comparative motion}
\label{fig:video}
\end{figure*}
Recent advancements in image generation~\cite{cao2024survey,zhang2023text,xu2024imagereward,ruiz2023dreambooth} and video generation~\cite{bar2024lumiere,chen2025jointtuner,kong2024hunyuanvideo} have demonstrated remarkable success, enabling a wide range of applications in computer graphics, cultural heritage, and the arts. Despite this substantial progress, particularly with diffusion-based models, the capability to generate videos with richer and more complex motion dynamics remains underexplored, especially in scenarios involving multi-character interaction. As illustrated in Fig.~\ref{fig:video}, prominent existing models such as 
WanX~\cite{wan2025wan}, face significant challenges in synthesizing smooth and rich interactive actions, particularly in long-term generation tasks.

To mitigate these challenges, various methods \cite{hu2024animate,peng2024conditionvideo,wang2024customvideo,zhang2024mimicmotion,tan2024AnimateX} have been developed that incorporate additional conditioning information, such as human motion priors, key frames, or reference videos, to enhance generation quality. Among these, using human motion data as a prior is a highly effective strategy to mitigate limb distortions, enhance motion expressiveness, and ensure long-term temporal coherence.

The increasing reliance on motion priors has, in turn, placed higher demands on motion generation models themselves. Current models predominantly operate in 3D space~\cite{jiang2023motiongpt,guo2024momask,zhang2024motiongpt,li2024lamp,tevet2023human,barquero2023belfusion,chen2023executing,zhang2024motiondiffuse,liang2024intergen}, mainly employing diffusion-based or autoregressive (e.g., GPT-based) frameworks. While many have achieved considerable success in generation quality and inference speed, a significant limitation is their predominant focus on single-character motion. Consequently, generated motions often lack diversity and fail to capture the nuances of interactions common in real-world scenarios. This limitation is largely attributed to the restricted scale and diversity of available training data. For instance, HumanML3D~\cite{guo2022generating}, a widely used dataset, contains only approximately 14,000 single-character motion clips. Although datasets for multi-character motion, like InterHuman~\cite{liang2024intergen} (with around 7,779 dual-character sequences), represent initial steps, they too suffer from limited scale and diversity. This data scarcity primarily stems from the inherent difficulty and high cost associated with capturing 3D motion data, which often necessitates expensive, high-precision motion capture systems.

2D motion representation offers a compelling alternative to address the data acquisition challenge, as annotations can be obtained more cost-effectively by leveraging advanced 2D pose estimation models~\cite{jiang2023rtmpose,khirodkar2024sapiens} and large language models (LLMs)~\cite{GPT4o,Gemini2Flash,ye2024mplug} for automated data processing from readily available videos. Furthermore, focusing on 2D motion is highly pertinent for image and video generation tasks, as it can simplify certain spatial complexities inherent in 3D representations while still providing sufficiently rich prior information. Recognizing this, recent efforts have aimed to curate large-scale 2D human motion datasets by mining vast quantities of internet videos. For example,~\cite{wang2024holistic} collected and annotated an extensive dataset of motion-caption pairs using pose estimation and LLMs, scaling 2D human motion data to one million pairs. Building on this,~\cite{wang2025humandreamer} introduced more fine-grained annotation and cleaning pipelines, proposing a dataset of 1.2 million motion-caption pairs for human pose generation. Despite these significant advancements in scale, a critical gap remains: these datasets still predominantly focus on single-character motions, largely overlooking the complexities of rich, multi-character interactive actions.

To address the identified gap in data availability for rich interactive motions, we introduce Motion2D-Video-150K, a novel large-scale dataset for 2D human motion generation. Motion2D-Video-150K comprises 150,000 motion sequences, specifically curated to cover a diverse range of both single-character and, crucially, double-character interactive motions, effectively leveraging vast Internet data.

Furthermore, building upon Motion2D-Video-150K, we propose RVHM2D, a novel human motion generation model. RVHM2D is designed to synthesize high-fidelity 2D human motion conditioned on both textual prompts and an initial motion frame. Notably, RVHM2D innovatively incorporates a reinforcement learning framework where the Fréchet Inception Distance (FID) serves as a reward signal to further enhance the quality and realism of the generated motions. The 2D motions generated by RVHM2D can then be seamlessly applied to drive video generation tasks using existing frameworks like skeleton-based ControlNet~\cite{zhao2024uni,zhang2023adding}, as exemplified in Fig.~\ref{fig:video}.

Our primary contributions are summarized as follows:
\begin{itemize}
\item We introduce Motion2D-Video-150K, a new large-scale 2D rich motion dataset containing 150,000 sequences. To the best of our knowledge, Motion2D-Video-150K is the first and largest publicly available dataset of its kind to comprehensively feature both single-character and complex double-character interactive motions.
\item We propose RVHM2D, a novel generative model for 2D human motion conditioned on both text prompts and an initial motion frame. RVHM2D effectively unifies the generation of single-character and double-character rich motions within a single, coherent framework and is capable of generating motions up to 300 frames.
\item We innovate by formulating the motion generation training process for RVHM2D as a reinforcement learning task, uniquely employing the Fréchet Inception Distance (FID) as a reward signal to guide the model towards generating higher-quality and more perceptually realistic motions.
\item We conduct comprehensive experiments demonstrating that RVHM2D, trained on Motion2D-Video-150K, achieves state-of-the-art performance against re-implemented baseline methods in the task of diverse and realistic human motion generation.
\end{itemize}

\begin{figure*}[t]
\begin{center}
\includegraphics[width=0.99\linewidth]{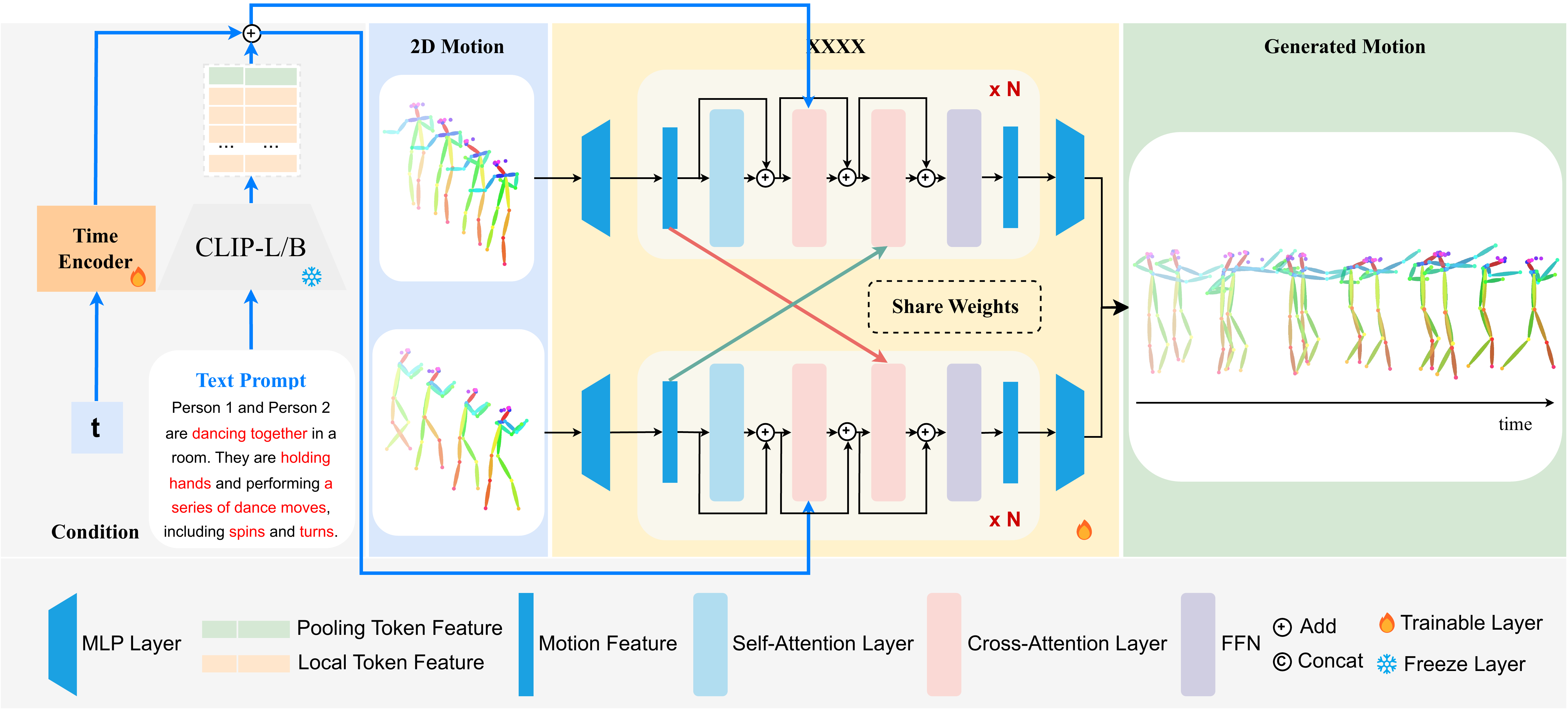}
\end{center}
\caption{The architecture of our proposed RVHM2D model for human motion generation.}
\label{fig:framework_RVHM2D}
\end{figure*}

\section{Related Work}
\subsection{Controllable Text-to-Video Generation}
\label{II-A}
Recent efforts in text-to-video generation have increasingly focused on controllability~\cite{bar2024lumiere, chen2025jointtuner, peng2024conditionvideo,wang2024customvideo}. Given that motion is a primary differentiator from static images, leveraging motion priors is a key strategy. For instance, ControlVideo~\cite{peng2024conditionvideo}, analogous to ControlNet~\cite{zhao2024uni} for images, introduced training-free control by incorporating cross-frame interactions into ControlNet's attention modules, improving video quality and consistency. Many other works~\cite{feng2023dreamoving, ma2024follow, zhai2024idol, li2025tokenmotion} have also adopted similar structures. Other recent works combine human motion with reference images to enhance character animation and robustness~\cite{hu2024animate, zhang2024mimicmotion, tan2024AnimateX}. However, despite these advancements, generating videos with complex, richly interactive multi-character scenarios remains a significant challenge, often limited by the expressiveness or availability of suitable motion priors.

\subsection{Text-to-Motion Generation}
\label{II-B}
Text-to-motion generation~\cite{sun2024lgtm,zhang2024motiondiffuse,shafir2023human,liang2024intergen,tevet2023human,jiang2023motiongpt,barquero2023belfusion} is a prominent task due to the intuitive nature of textual input. Early efforts often focused on 3D single-character human motion, with main approaches including LLM-based methods~\cite{jiang2023motiongpt,guo2024momask,zhang2024motiongpt,li2024lamp} that autoregressively generate tokenized motion, and diffusion-based methods~\cite{tevet2023human,barquero2023belfusion,chen2023executing,zhang2024motiondiffuse,liang2024intergen} that learn a denoising process, sometimes also using VQVAE for discretization. While recent advancements like MotionLCM~\cite{dai2024motionlcm} and Motion Mamba~\cite{zhang2024motion} have improved single-character motion quality and efficiency, a persistent limitation, even in these advanced models, is the generation of rich, interactive motions involving multiple characters.

Specific attempts at multi-character motion generation exist. For example, priorMDM~\cite{jiang2023motiongpt} extended MDM for two-person scenarios using an additional Transformer, InterGen~\cite{liang2024intergen} redesigned the diffusion process leveraging interaction symmetries, and MoMat-MoGen~\cite{cai2024digital} employed a retrieval-based strategy for priors. Our proposed RVHM2D model also adopts a diffusion-based approach but aims to unify high-quality single- and multi-character (specifically, double-character) motion generation within a single framework, conditioned on rich textual prompts and benefiting from our new Motion2D-Video-150K dataset.

Despite these efforts, existing multi-character motion generation still faces critical hurdles: 1) a significant scarcity of large-scale, diverse training data for complex interactions (which our Motion2D-Video-150K dataset aims to alleviate); 2) limited realism and complexity in the generated interactions; and 3) difficulties in precise semantic and detailed control of these interactions.

\begin{figure*}[t]
\centering
\includegraphics[width=0.99\linewidth]{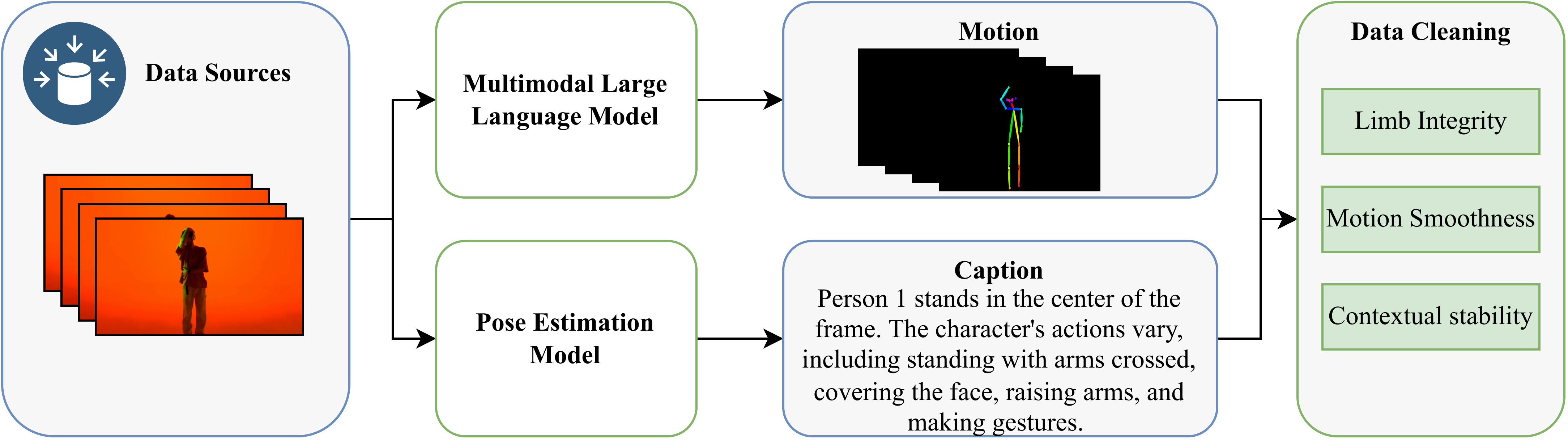}
\caption{The annotation and data cleaning pipeline for our Motion2D-Video-150K human motion 2D dataset. This pipeline involves initial pose and text annotation followed by rigorous filtering based on limb integrity, motion smoothness, and contextual stability.}
\label{fig:dataclean_pipeline}
\end{figure*}

\subsection{Reinforcement Learning for Generative Modeling}
\label{II-C}
Reinforcement Learning (RL) provides a paradigm for optimizing objectives through interaction. An MDP, defined by $(S, A, P, R, \gamma)$, aims to find a policy $\pi(a|s)$ maximizing cumulative reward. In generative modeling, RL can optimize non-differentiable metrics or fine-tune models. Numerous studies~\cite{wallace2024diffusion, ethayarajh2024kto, wang2024rlcoder, cideron2024musicrl, collins2024beyond} have employed RL to enhance model performance across various tasks. However, its application to human motion synthesis, particularly for directly enhancing perceptual quality using metrics like Fréchet Inception Distance (FID), remains relatively underexplored. Our work explores the integration of an FID-based objective to further refine the generation quality of our RVHM2D model.

\section{Method}
\label{sec:method}
In this section, we first detail the collection, annotation, and cleaning pipeline for our proposed Motion2D-Video-150K dataset. Subsequently, we will present the methodology underpinning our model, RVHM2D.

\subsection{The Motion2D-Video-150K Dataset: A 150K Rich Video Human-Motion2D Dataset}
\label{subsec:dataset}
Addressing the limitations of existing datasets in capturing diverse and interactive multi-character motions, as discussed in ~\ref{II-B}, we construct Motion2D-Video-150K, a large-scale 2D rich motion dataset.

\textbf{Data Sources:}
The Motion2D-Video-150K dataset is curated from two primary sources to ensure diversity in motion and character interactions. Firstly, we incorporate data from established open-source human motion datasets, including HAA500~\cite{chung2021haa500}, Penn Action Dataset~\cite{zhang2013actemes}, and UCF101~\cite{soomro2012ucf101}. These datasets predominantly feature single-character, human-centric videos, which are valuable for learning fundamental human skeletal structures and movements. Secondly, to gather a rich collection of double-character interactions, we collected over 500,000 video clips from online platforms. The search queries for this collection were generated using GPT-4o and included terms such as \texttt{"group work"} and \texttt{"cooperation"}, with a focus on capturing two-character interactions.

\textbf{Data Annotation:}
Our Motion2D-Video-150K dataset consists of 2D motion sequences paired with textual descriptions. We define a video sample in our dataset, $V$, as containing one or more character motion sequences. Each individual character's 2D skeleton sequence, $s_c = \{k_j\}_{j=1}^{L_c}$, comprises $L_c$ frames, where $k_j \in \mathbb{R}^{N \times 3}$ represents the $N=17$ keypoints at frame $j$. Each keypoint is defined by its x-coordinate, y-coordinate, and a confidence score. Each video sample $V$ is also associated with a textual description $c_t$.

For motion annotation, we employed RTMPose-Large~\cite{jiang2023rtmpose}, a robust model that integrates human bounding box detection and subsequent skeleton keypoint estimation. Concurrently, for textual annotation, we utilized the Gemini 2.0 Flash model~\cite{Gemini2Flash} as well as Owl3 model ~\cite{ye2024mplug} with the following prompt designed to elicit detailed descriptions of actions, interactions, and spatial relationships:

\texttt{"You are an AI specialized in analyzing video sequences. Given a series of images from the same video clip, describe the actions, interactions, and positional relationships of the characters in a single sentence, treating the images as a continuous sequence. Follow these requirements: \\ 1. Number the characters from left to right based on their positions in the first image, and consistently refer to each character by the same number throughout. \\ 2. Describe the characters' actions and their left-to-right positional relationships as they change across the sequence. \\ 3. Use a sentence structure starting with 'Person 1 ...' \\ 4. Avoid mentioning the characters' clothing or the video's background. \\ 5. Do not describe each image individually or include introductory phrases like 'Here is a description of the video.'"
}

\textbf{Data Cleaning:}
Initial annotations inevitably contained noisy samples, such as sequences with incomplete bodies, inconsistent numbers of tracked persons, overly erratic movements, or pose misdetections. To ensure data quality, we developed a multi-stage cleaning pipeline, illustrated in Fig.~\ref{fig:dataclean_pipeline}, focusing on three key aspects:

\begin{figure*}[t]
\centering
\includegraphics[width=0.99\linewidth]{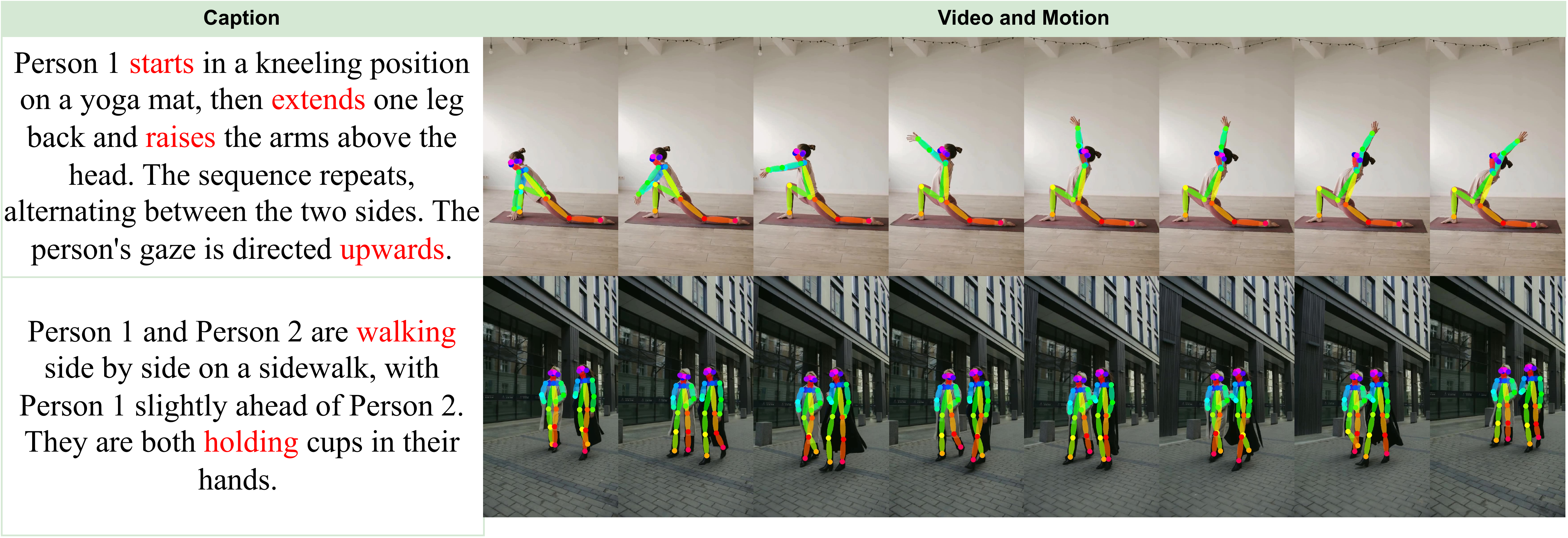}
\caption{An example from the Motion2D-Video-150K 2D rich motion dataset. The textual caption is generated by Gemini 2.0 Flash, and the 2D skeletons are extracted using RTMPose-Large.}
\label{fig:dataset_example}
\end{figure*}

\begin{itemize}
\item \textbf{Limb Integrity:} A character is considered valid if the average confidence score across all 17 keypoints is above 0.5 and the average confidence score for facial keypoints also exceeds 0.5.
\item \textbf{Motion Smoothness:} Assuming that human motion in videos should be relatively smooth and continuous, we penalize overly erratic movements. For each valid character, we calculate the mean inter-frame displacement of their keypoints. Let $k_{f,p}$ be the pose vector for person $p$ at frame $f$. We compute a frame-wise difference, e.g., $d_f = \frac{1}{N} \sum_{i=1}^{N} ||k_{f,p,i} - k_{f-1,p,i}||_2$. If $d_f$ frequently exceeds a predefined threshold $\tau_{smooth}$, the video segment is flagged. Videos with an excessive number of such irregular movements are discarded.
\item \textbf{Contextual Stability:} We monitor significant fluctuations in the number of validly tracked persons throughout a video. Sequences exhibiting unstable tracking or frequent changes in the number of interacting characters are removed to maintain contextual consistency.
\end{itemize}

\textbf{Dataset Analysis and Statistics:}
Following the rigorous cleaning process, the Motion2D-Video-150K dataset comprises 150,000 high-quality 2D rich motion sequences. A key characteristic of Motion2D-Video-150K is its emphasis on interactive scenarios; the ratio of video segments containing two-person motions to those with single-person motions is approximately 1.5:1. In terms of temporal length, 
the dataset contains sequences of varying durations, with many extending up to 300 frames, which also defines the maximum sequence length processed by our proposed model RVHM2D. Motion2D-Video-150K spans over 300 distinct motion categories, encompassing a wide array of both single-character and intricate double-character interactions. 
An example illustrating the annotated 2D skeletons and the corresponding generated caption is provided in Fig.~\ref{fig:dataset_example}.

\textbf{Data Splits and Availability:}
The complete Motion2D-Video-150K dataset comprises 150,000 motion sequences. For evaluation purposes, we randomly sampled 11,682 sequences to constitute the test set. This test set is carefully balanced to reflect the diverse nature of our dataset, containing 6,260 single-character motion sequences and 5,422 double-character (interactive) motion sequences. These sequences are held out and used exclusively for final performance reporting. All the remaining data are used to train out model.

\begin{figure*}
\centering
\includegraphics[width=0.99\linewidth]{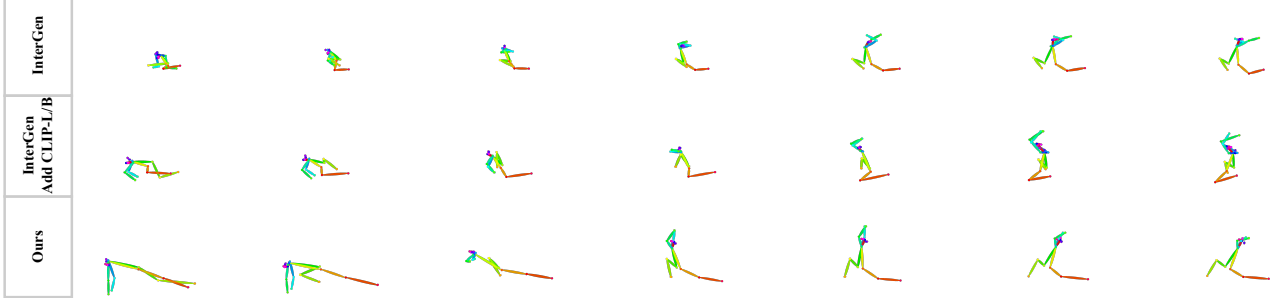}
\caption{Qualitative comparison of generated 2D human motions from different models: (a) InterGen, (b) InterGen with enhanced text encoders (CLIP-L/B), and (c) Our full RVHM2D model.}
\label{fig:qualitative_comparison}
\end{figure*}

\subsection{RVHM2D: A Diffusion-based Rich Motion Generation Model}
\label{sec:model_RVHM2D}
Our proposed model, RVHM2D, is designed to generate rich 2D human motions, including complex single- and double-character interactions, conditioned on textual prompts and optionally an initial reference motion frame. RVHM2D integrates several key components: advanced textual feature utilization, a reinforcement learning training method using Fréchet Inception Distance (FID), and a specialized diffusion model architecture.

\textbf{Model Architecture:}
Similar to InterGen~\cite{liang2024intergen}, RVHM2D adopts a dual-tower structure to ensure the stability of the interaction effect between double characters. In order to integrate both single and double characters scenarios into our model, we add \texttt{"1"} or \texttt{"2"} into the text prompt to indicate the number of people in current case, and specifically for the single-character cases, we replicate the same skeleton twice to simulate a double-character interaction. Additionally, we utilize CLIP-L and CLIP-B text encoders together in a concatenating manner where each piece of text input will go through both text encoders and the encoder outputs will be concatenated together to obtain the text features and we also tried utilizing T5-XXL text encoder. We retain their final global features and local token features. To accommodate text prompts exceeding standard token limit and preserve all textual information, we divide the tokens into several sequences, and the input sequences are separately sent into our text encoder(s) for processing. After that, the out features are concatenated again to ensure no text information is lost. Then, combining the DDIM~\cite{song2020denoising} sampler, we randomly choose a time-step $t$ and add noise into the input skeleton points. Simultaneously, the $t$ is encoded by sinusoidal function and $2$ layers of $MLP$, obtaining features $f_t$. Then, $f_t$ is injected into the de-noising process of the 2D motion generation. In addition, we use transformer-based decoders module with 8 layers to extract motion features, similar to the UNet structure in SD, and inject text features at each layer. Next, we input the two motion into the two branches of the shared parameters respectively, and inject the text features into the subsequent single-character features. Furthermore, we can calculate the interactive attention~\cite{liang2024intergen} and the specific calculation process is as follows:
$$ h_1 = SA(a,c_{t}) + a$$
$$ h_2 = CA_1(h_1,f_{local}) + h_1$$
$$ h_3 = CA_2(h_2, b, c_{t}) + h_2$$
where $a$ and $b$ refer to the input noised motion embeddings of the two person. $SA$ and $CA$ represent self-attention layer and cross-attention layer separately. $c_{t}$ is calculated by adding text pooling feature $f_{pooling}$ and time-step feature $f_t$. The overall framework is depicted in Fig.~\ref{fig:framework_RVHM2D}. 

RVHM2D also supports conditioning on an initial motion frame $\mathbf{m}_{start}$ to guide the generation. This reference frame is encoded using an MLP to obtain a feature representation $\mathbf{f}_{ref\_motion}$. This feature is then integrated into the denoising network after the initial self-attention layer of each Transformer block using an additional self-attention mechanism over the concatenated sequence and reference features:
\begin{equation}
\begin{split}
\mathbf{h}_{\text{refined}} &= \mathrm{SA}_{\text{ref}}(\mathbf{h}_{\text{1}}, \mathbf{f}_{\text{ref\_motion}}) \\
&= \mathrm{Attention}\bigl(Q = \mathbf{h}_{\text{1}},\; K=V=[\mathbf{h}_{\text{1}}, \mathbf{f}_{\text{ref\_motion}}]\bigr)
\end{split}
\label{eq:ref_motion_attn}
\end{equation}

\textbf{Diffusion Model:} Diffusion~\cite{song2020denoising,ho2020denoising} is modeled as a Markov noising process~\cite{dynkin1965markov}, $\{s^t\}_{t=0}^T$, where $s^0$ is derived from the data distribution and
\begin{equation}
  q(s^t|s^{t-1}) = \mathcal{N}( \sqrt{\alpha_t}s^{t-1}, (1 - \alpha_t)I)
  \label{eq:eq1}
\end{equation}
where $\alpha_t\in(0,1)$ are constant hyper-parameters. If $\alpha_t$ is sufficiently small, $s^T$ can be approximated as $s^T\sim\mathcal{N}(0,I)$. In addition, we define $t$ as the noising step. 
In our approach, with only considering text condition $c_t$, the conditioned motion synthesis models the distribution $p(s^0|c_t)$ as the reversed diffusion process of gradually refining $s^T$. Instead of predicting $\epsilon_t$ as formulated by \cite{ho2020denoising}, we predict the signal itself~\cite{ramesh2022hierarchical} $\hat{s}^0 = G(s^t, t, c_t)$ with the objective:
\begin{equation}
\mathcal{L}_{\text{simple}} = \mathbb{E}_{s^0 \sim q(s^0|c_t), t \sim [1, T]} \left[ \| s^0 - G(s^t, t, c_t) \|_2^2 \right]
\label{eq:eq2}
\end{equation}

\begin{table*}[t]
\centering
\small
\setlength{\tabcolsep}{4.35pt}
\begin{tabular}{c|c|c|c|c|c|c|c|c|c|c|c|c}
\hline
\multirow{3}{*}{Method}& \multicolumn{6}{|c|}{Single Character}& \multicolumn{6}{|c}{Two Characters }\\
\cline{2-13}
\multirow{3}{*}{}& \multicolumn{3}{|c|}{R Precision $\uparrow$}& \multirow{2}{*}{FID$\downarrow$}& \multirow{2}{*}{MM Dist$\downarrow$}& \multirow{2}{*}{Diversity$\rightarrow$}& \multicolumn{3}{|c|}{R Precision$\uparrow$} & \multirow{2}{*}{FID$\downarrow$}& \multirow{2}{*}{MM Dist$\downarrow$} & \multirow{2}{*}{Diversity$\rightarrow$}\\
\cline{2-4}\cline{8-10}
& Top1& Top2& Top3& & & & Top1& Top2& Top3& && \\
\hline
MDM & 23.0& 36.84& 47.02& \textbf{0.2674}& 1.0092& 1.0220& 27.74& 43.24& 54.0& \textbf{0.3926} & 0.9775& 1.1644\\
\hline
InterGen & 33.67& \textbf{51.09} & \textbf{60.50} & 1.4317 & \textbf{0.7721} & 3.6571 & 30.70 & \textbf{45.70} & \textbf{55.94} & 2.0633 & \textbf{0.8220} & 3.6201 \\
\hline
Ours (RVHM2D) & \textbf{36.64}& 48.83& 56.92 & 1.9241 & 0.8021 & \textbf{3.7617} & \textbf{31.48}& 45.31& 54.84 & 2.4123 & 0.8497 & \textbf{3.7574} \\
\hline
\end{tabular}
\caption{Quantitative comparison with state-of-the-art methods on the Motion2D-Video-150K test set for both single-character and two-character 2D motion generation. Best results are in \textbf{bold}. $\uparrow$ indicates higher is better, $\downarrow$ lower is better, $\rightarrow$ indicates closer to ground truth human motion diversity is better.}
\label{tab:sota_comparison}
\end{table*}

\textbf{Enhanced Textual Conditioning:}
Conventional text-to-motion methods often rely solely on global pooled features extracted from text encoders. However, for generating nuanced and complex motions, particularly those involving interactions described in detail by our Motion2D-Video-150K dataset's captions, such global features can be insufficient. Therefore, inspired by~\cite{yu2023zero}, we augment global textual features with fine-grained local features extracted from the CLIP-L/B text encoder or T5-XXL text encoder, capturing word-level semantic information. These local features are then injected into our model's decoder via an attention mechanism to provide more detailed guidance.

\textbf{Reinforcement Learning-Inspired Refinement:}
Inspired by successes in other generative domains (e.g. large language models~\cite{GPT4o}), we explore a refinement strategy for RVHM2D that draws from reinforcement learning (RL) principles to align generated motions more closely with desired perceptual qualities. While RL has been applied in various generative tasks, its use for refining human motion generation with comprehensive perceptual metrics like FID as a direct optimization target remains relatively underexplored.

We formulate the motion generation task where the state $s$ corresponds to the input text prompt $c_t$, and the action $a$ is the generated motion sequence $m$. The generated motion sequence $m$ is defined as:
\begin{equation}
m = \{ \mathbf{m}_1, \mathbf{m}_2, \ldots, \mathbf{m}_T \}
\label{eq:motion_sequence}
\end{equation}
where $T$ is the total number of frames. Each frame $\mathbf{m}_f$ (for $f=1, \ldots, T$) contains the poses of the characters present. For a two-character scenario, common in our Motion2D-Video-150K dataset, we define:
\begin{equation}
\mathbf{m}_f = \{\mathbf{p}_{f,1}, \mathbf{p}_{f,2}\}
\label{eq:frame_poses}
\end{equation}
where $\mathbf{p}_{f,c}$ is the pose of character $c$ at frame $f$. Each character's pose is represented by $N=17$ keypoints:
\begin{equation}
\mathbf{p}_{f,c} = \{\mathbf{k}_{j,c}\}_{j=1}^N
\label{eq:pose_keypoints}
\end{equation}
where $\mathbf{k}_{j,c} \in \mathbb{R}^3$ includes the x-coordinate, y-coordinate, and confidence score for the $j$-th keypoint of character $c$. The policy $\pi(m|c_t)$ is embodied by our generative model RVHM2D. 

What is worth mentioning is that we devise a two-stage training strategy: the model is first trained with a standard diffusion objective, and then fine-tuned using reinforcement learning with an FID-based reward to further enhance motion realism and text alignment. The weight for the reward signal is set very low to stabilize training and prevent policy collapse. In the second stage, we incorporate FID-based scores directly into our loss function 
to serve as a strong learning signal, akin to a reward in RL, guiding the model towards perceptually superior outputs.

\subsection{Loss Function}
\label{sec:loss_function}
Motion generation introduces temporal consistency challenges that necessitate specialized loss functions. Inspired by InterGen~\cite{liang2024intergen}, our model retains bone length loss, velocity loss, distance map loss, joint awareness loss and reconstruction loss. In the first training stage, we train our model directly with these losses, where $\lambda_{(\cdot)}$ are hyperparameter weights for each loss term.
\begin{equation}
\begin{split}
\mathcal{L}_{\text{first\_stage}} ={}& \lambda_{BL}\mathcal{L}_{BL} + \lambda_{VEL}\mathcal{L}_{VEL} + \lambda_{DM}\mathcal{L}_{DM} + \\&\lambda_{JA}\mathcal{L}_{JA} + \lambda_{Recon}\mathcal{L}_{Recon}
\end{split}
\label{eq:loss_total}
\end{equation}

In the second stage, to enhance perceptual quality and alignment with data distributions, we incorporate losses derived from pre-trained Fréchet Inception Distance (FID) evaluation models, trained following~\cite{guo2022generating}. Unlike L1/L2 losses, FID captures higher-level distributional similarities. This component includes:
\begin{itemize}
\item Text-Motion FID Loss ($\mathcal{L}_{\text{textfid}}$): Measures the similarity between the generated motion and the input text prompt $c_t$.
\begin{equation}
\mathcal{L}_{\text{textfid}} = 1 - \text{sim}_{\text{cos}}(\text{Enc}_{\text{text}}(c_t), \text{Enc}_{\text{motion}}(\mathbf{m}_{pred}))
\label{eq:loss_textfid}
\end{equation}
where $\text{Enc}_{\text{text}}(\cdot)$ and $\text{Enc}_{\text{motion}}(\cdot)$ are pre-trained encoders from the FID evaluation model, and $\text{sim}_{\text{cos}}(\cdot,\cdot)$ is cosine similarity.
\item Motion FID Loss ($\mathcal{L}_{\text{motionfid}}$): Measures the similarity between the generated motion and the ground truth motion distribution.
\begin{equation}
\mathcal{L}_{\text{motionfid}} = 1 - \text{sim}_{\text{cos}}(\text{Enc}_{\text{motion}}(\mathbf{m}_{gt}), \text{Enc}_{\text{motion}}(\mathbf{m}_{pred}))
\label{eq:loss_motionfid}
\end{equation}
\end{itemize}

The final objective is a weighted sum of these individual loss components:
\begin{equation}
\begin{split}
\mathcal{L}_{\text{total}} =\mathcal{L}_{\text{first\_stage}}+ \lambda_{\text{textfid}}\mathcal{L}_{\text{textfid}} + \lambda_{\text{motionfid}}\mathcal{L}_{\text{motionfid}}
\end{split}
\label{eq:loss_total}
\end{equation}
where $\lambda_{(\cdot)}$ are hyperparameter weights for each loss term.

\section{Experiments}

\begin{table*}[t]
\centering
\small
\setlength{\tabcolsep}{3pt}
\begin{tabular}{c|c|c|c|c|c|c|c|c|c|c|c|c|c|c}
\hline
\multirow{3}{*}{\shortstack{(CLIP-L/B)}}& \multirow{3}{*}{\shortstack{Text \\ FID}}& \multirow{3}{*}{\shortstack{Motion \\ FID}}& \multicolumn{6}{|c|}{Single Character}& \multicolumn{6}{|c}{Two Characters }\\
\cline{4-15}
\multirow{3}{*}{}& \multirow{3}{*}{}& \multirow{3}{*}{}& \multicolumn{3}{|c|}{R Precision$\uparrow$}& \multirow{2}{*}{FID$\downarrow$}& \multirow{2}{*}{MM Dist$\downarrow$}& \multirow{2}{*}{Diversity$\rightarrow$}& \multicolumn{3}{|c|}{R Precision$\uparrow$} & \multirow{2}{*}{FID$\downarrow$}& \multirow{2}{*}{MM Dist$\downarrow$} & \multirow{2}{*}{Diversity$\rightarrow$}\\
\cline{4-6}\cline{10-12}
& & & Top1& Top2& Top3& & & & Top1& Top2& Top3& && \\
\hline
 & & & 33.67& 51.09 & \textbf{60.50} & 1.4317 & 0.7721 & 3.6571 & 30.70 & 45.70 & 55.94 & 2.0633 & \textbf{0.8220} & 3.6201 \\
\hline
\checkmark& & & 34.14 & \textbf{51.48} & 60.00& \textbf{1.3931} & \textbf{0.7714} & 3.6881 & 30.70 & 45.16& \textbf{56.02} & \textbf{2.0268} & 0.8230 & 3.6060 \\
\hline
\checkmark & \checkmark&& 34.16 & 48.55 & 57.47 & 1.7835 & 0.7999 & 3.7505 & 30.78 & 45.62 & 55.31 & 2.3520 & 0.8470 & 3.5747 \\
\hline
\checkmark && \checkmark& 34.81 & 48.17 & 56.75& 1.7986 & 0.8000 & 3.7320 & 31.17 & \textbf{46.56} & 55.70 & 2.3915 & 0.8474 & 3.6346 \\
\hline
\checkmark & \checkmark& \checkmark& \textbf{36.64} & 48.83 & 56.92 & 1.9241 & 0.8021 & \textbf{3.7617} & \textbf{31.48} & 45.31 & 54.84 & 2.4123 & 0.8497 & \textbf{3.7574} \\
\hline
\end{tabular}
\caption{Ablation study on key components of the RVHM2D framework. The first row refers to InterGen as baseline. Subsequent rows incrementally add our proposed text encoding enhancements (CLIP-L/B + Local Features), Text FID loss, and Motion FID loss.}
\label{tab:ablation_components}
\end{table*}

\begin{table*}[t]
\centering
\small
\setlength{\tabcolsep}{4.7pt}
\begin{tabular}{c|c|c|c|c|c|c|c|c|c|c|c|c}
\hline
\multirow{3}{*}{\shortstack{Text Encoder}} & \multicolumn{6}{c|}{Single Character} & \multicolumn{6}{c}{Two Characters}\\
\cline{2-13} %

& \multicolumn{3}{c|}{R Precision$\uparrow$} & \multirow{2}{*}{FID$\downarrow$} & \multirow{2}{*}{MM Dist$\downarrow$} & \multirow{2}{*}{Diversity$\rightarrow$} & \multicolumn{3}{c|}{R Precision$\uparrow$} & \multirow{2}{*}{FID$\downarrow$} & \multirow{2}{*}{MM Dist$\downarrow$} & \multirow{2}{*}{Diversity$\rightarrow$}\\
\cline{2-4} \cline{8-10} %
& Top1 & Top2 & Top3 & & & & Top1 & Top2 & Top3 & & & \\
\hline
CLIP-L/B & 36.64 & 48.83 & 56.92 & 1.9241 & 0.8021 & 3.7617 & 31.48 & 45.31 & 54.84 & 2.4123 & 0.8497 & \textbf{3.7574} \\
\hline
T5-XXL& \textbf{40.58}& \textbf{53.97}& \textbf{62.59} & \textbf{0.6516} & \textbf{0.7123} & \textbf{3.9384} & \textbf{34.81}& \textbf{48.97} & \textbf{58.52} & \textbf{1.0153} & \textbf{0.7872} & 3.7573 \\
\hline
\end{tabular}
\caption{Ablation study on the text encoder(s).}
\label{tab:ablation_text_encoder}
\end{table*}

\subsection{Experimental Setup}

\textbf{Implementation Details:}
We re-implemented two %
strong baseline models for the 2D human motion generation task: 
MDM~\cite{tevet2023human}
and InterGen~\cite{liang2024intergen}. All models, including our proposed RVHM2D model and the re-implemented baselines, were trained using the AdamW optimizer with a fixed learning rate of $1 \times 10^{-4}$. Training was performed on 8 NVIDIA A100 GPUs for 300 epochs. The batch size was set to 32 or 16 per GPU, depending on the specific memory requirements of each model. For our RVHM2D model and the diffusion-based baselines, the number of diffusion steps during training was set to 1000. For models that incorporate reinforcement learning, the training was divided into two stages: 200 epochs of initial training followed by 100 epochs of RL-based fine-tuning. Inference settings, including the number of steps for baselines, are kept consistent with their original implementations for fair comparison.

\textbf{Evaluation Metrics:}
Following established practices in 3D human motion generation~\cite{guo2022generating}, we adopt a comprehensive set of metrics to evaluate performance: R-Precision (Top-1, Top-2, Top-3), Fréchet Inception Distance (FID), Multimodal Distance (MM Dist), and Diversity. 
To extract features for these metrics, we trained an evaluation model whose architecture is similar to that proposed in InterGen~\cite{liang2024intergen}. However, to align with the text encoding capabilities of our RVHM2D model, our evaluation feature extractor utilizes CLIP-L/B text encoders as our proposed model.

\subsection{Quantitative Comparisons}

\textbf{Comparison with State-of-the-Art Methods:}
We compare our proposed RVHM2D model with the re-implemented baselines (MDM, 
and InterGen) on the Motion2D-Video-150K test set. We extend MDM to both single-character and double-character scenarios, by duplicating the single-character data.

In contrast to the baselines, our RVHM2D model is designed to inherently handle both single- and double-character generation, leveraging enhanced text feature utilization and an FID-based refinement strategy. The comparative results are presented in Table~\ref{tab:sota_comparison}.

As shown in Table~\ref{tab:sota_comparison}, our RVHM2D model demonstrates strong performance. Specifically, for single-character generation, RVHM2D achieves an R-Precision-Top1 of 36.64, surpassing InterGen (33.67). For two-character generation, RVHM2D obtains an R-Precision-Top1 of 31.48, also outperforming InterGen (30.70). While our proposed model performs slightly behind on R-Precision in Top2 and Top3 and MM Dist, our RVHM2D model shows superior R-Precision in Top1 for both single and two-character scenarios and leads in Diversity, indicating its capability to generate varied and text-relevant motions. MDM achieves the best FID scores, but falls behind on other metrics, indicating its over fitting problem.

\subsection{Ablation Studies}
We conduct comprehensive ablation studies to validate the effectiveness of different components within our RVHM2D model and investigate the impact of various design choices.

\textbf{Impact of Components:}%
To demonstrate the effects of our dual CLIP-L/B text encoder configuration, the utilization of local text features, and our FID-based refinement strategy, we performed several ablation experiments. The results are presented in Table~\ref{tab:ablation_components}. We start with InterGen as our initial baseline, as reported in Table~\ref{tab:sota_comparison} and incrementally add our proposed enhancements.

As shown in Table~\ref{tab:ablation_components}, by using CLIP-L/B text encoders and leveraging local text features, we improve R-precision-Top1 by 0.47\%, FID by 0.0386, MM distance by 0.0007 and diversity by 0.031 in single character motion generation, as well as R-precision-Top3 by 0.08\% and MM distance by 0.0365. Furthermore, incorporating the Text FID and Motion FID loss components contributes to R-Precision-Top1 and Diversity with slight falling on other metrics, suggesting that this FID-based refinement helps generate motions that are more consistent with textual descriptions and closer to the distribution of real motions. After that, we incorporate all the modules to get our full model architecture. 

\begin{table*}[t]
\centering
\small
\setlength{\tabcolsep}{4.5pt}
\begin{tabular}{c|c|c|c|c|c|c|c|c|c|c|c|c}
\hline
\multirow{3}{*}{\begin{tabular}{@{}c@{}}Method \\ Configuration\end{tabular}} & \multicolumn{6}{|c|}{Single Character}& \multicolumn{6}{|c}{Two Characters }\\
\cline{2-13}
\multirow{3}{*}{}& \multicolumn{3}{|c|}{R Precision $\uparrow$}& \multirow{2}{*}{FID$\downarrow$}& \multirow{2}{*}{MM Dist$\downarrow$}& \multirow{2}{*}{Diversity$\rightarrow$}& \multicolumn{3}{|c|}{R Precision$\uparrow$} & \multirow{2}{*}{FID$\downarrow$}& \multirow{2}{*}{MM Dist$\downarrow$} & \multirow{2}{*}{Diversity$\rightarrow$}\\
\cline{2-4}\cline{8-10}
& Top1& Top2& Top3& & & & Top1& Top2& Top3& && \\
\hline
\begin{tabular}{@{}c@{}}Owl3 \\ w/o ref motion\end{tabular} & \textbf{36.64}& \textbf{48.83} & \textbf{56.92} & \textbf{1.9241} & \textbf{0.8021} & \textbf{3.7617} & 31.48 & 45.31& 54.84 & \textbf{2.4123} & 0.8497 & 3.7574\\
\hline
\begin{tabular}{@{}c@{}}Gemini \\ w/o ref motion\end{tabular} & 33.39 & 45.67 & 54.69 & 3.6463 & 0.8215 & 3.8016 & 32.86 & \textbf{46.55} & 55.14 & 3.8468 & \textbf{0.8265} & 3.7418 \\
\hline
\begin{tabular}{@{}c@{}}Gemini \\ w/ ref motion\end{tabular} & 32.86 & 45.78 & 54.48 & 3.7103 & 0.8246 & 3.8651 & \textbf{33.17} & 46.34 & \textbf{55.53} & 3.9059 & 0.8287 & \textbf{3.8411} \\
\hline
\end{tabular}
\caption{Ablation study on the impact of different text caption sources (Owl3 vs. Gemini 2.0 Flash) and the inclusion of first-frame reference motion for the RVHM2D model.}
\label{tab:ablation_caption_ref}
\end{table*}

\textbf{Impact of Different Text Encoder:}
Notably, when we replace the CLIP-based text encoders in our full RVHM2D architecture with a more powerful T5-XXL encoder~\cite{raffel2020exploring}, we observe a significant performance boost across most metrics, particularly FID and R-Precision. This underscores the importance of a strong text encoder for rich motion generation and demonstrates the capability of our RVHM2D architecture to leverage more powerful text features effectively. The results are shown in Table ~\ref{tab:ablation_text_encoder}

\textbf{Impact of Caption Source and Reference Motion:}
We also investigated the effect of different text caption sources and the inclusion of a first-frame reference motion. These results are presented in Table~\ref{tab:ablation_caption_ref}.

Observations from Table~\ref{tab:ablation_caption_ref} indicate that text annotations from Gemini 2.0 Flash, compared to Owl3, yield slightly lower R-Precision for single-character generation but can be beneficial for two-character generation, particularly improving R-Precision-Top2. Furthermore, when using Gemini-annotated text, incorporating the first frame as an additional motion prior effectively enhances R-Precision for two-character generation, with R-Precision-Top1 from 32.86 to 33.17.

\subsection{Qualitative Analysis}
\label{sec:qualitative_analysis}
To provide a visual assessment of generation quality, we present qualitative examples of 2D human motions generated by our RVHM2D model and compared baselines in Figure~\ref{fig:qualitative_comparison}.

As illustrated in Figure~\ref{fig:qualitative_comparison}, when comparing the baseline InterGen with a version enhanced by stronger CLIP-L/B text encoders, it is evident that improved text encoding enhances the semantic alignment between the input text and the generated motion. The baseline InterGen struggles with prompts, whereas the enhanced version more accurately captures these actions. Furthermore, our full RVHM2D model, benefiting from its comprehensive design including the FID-based refinement, demonstrates an ability to generate motions with more realistic human body structures and more natural, human-like dynamics compared to the other approaches.

\subsection{Downstream Application: Motion2D-Driven Video Synthesis}
To validate the practical applicability and assess the perceptual quality of motions generated by our RVHM2D model, we conducted experiments on skeleton-driven video synthesis. We leveraged Wan2.1-T2v-14B for pose-guided video generation. The 2D skeleton sequences $\mathbf{m}_{pred}$ produced by RVHM2D, conditioned on textual prompts served as the primary control input for the video synthesis model. Qualitative results are presented in Figure~\ref{fig:video}. The synthesized videos demonstrate a higher degree of motion realism and a more plausible execution of described actions.This suggests that the structural and temporal properties captured by RVHM2D are robust enough for downstream video applications, further highlighting the effectiveness of our approach in generating high-fidelity human motions.

\section{Conclusion and Future Work}
This paper introduced RVHM2D, a novel human motion generation model utilizing enhanced textual conditioning and an FID-based reinforcement learning strategy. We also presented Motion2D-Video-150K, a new large-scale 2D rich motion dataset featuring a balanced distribution of single- and double-character sequences. Experiments confirmed RVHM2D's effectiveness in producing high-fidelity, text-coherent motions. Current limitations include training efficiency and stability for complex long sequences, occasionally leading to positional floating and other artifacts. Future work will focus on accelerating training and improving the robustness and coherence of generating extended, intricate motions.


\begin{thebibliography}{55}
    \providecommand{\natexlab}[1]{#1}
    
    \bibitem[{Bar-Tal et~al.(2024)Bar-Tal, Chefer, Tov, Herrmann, Paiss, Zada, Ephrat, Hur, Liu, Raj et~al.}]{bar2024lumiere}
    Bar-Tal, O.; Chefer, H.; Tov, O.; Herrmann, C.; Paiss, R.; Zada, S.; Ephrat, A.; Hur, J.; Liu, G.; Raj, A.; et~al. 2024.
    \newblock Lumiere: A space-time diffusion model for video generation.
    \newblock In \emph{SIGGRAPH Asia 2024 Conference Papers}, 1--11.
    
    \bibitem[{Barquero, Escalera, and Palmero(2023)}]{barquero2023belfusion}
    Barquero, G.; Escalera, S.; and Palmero, C. 2023.
    \newblock Belfusion: Latent diffusion for behavior-driven human motion prediction.
    \newblock In \emph{Proceedings of the IEEE/CVF International Conference on Computer Vision}, 2317--2327.
    
    \bibitem[{Cai et~al.(2024)Cai, Jiang, Qing, Guo, Zhang, Lin, Mei, Wei, Wang, Yin et~al.}]{cai2024digital}
    Cai, Z.; Jiang, J.; Qing, Z.; Guo, X.; Zhang, M.; Lin, Z.; Mei, H.; Wei, C.; Wang, R.; Yin, W.; et~al. 2024.
    \newblock Digital life project: Autonomous 3d characters with social intelligence.
    \newblock In \emph{Proceedings of the IEEE/CVF Conference on Computer Vision and Pattern Recognition}, 582--592.
    
    \bibitem[{Cao et~al.(2024)Cao, Tan, Gao, Xu, Chen, Heng, and Li}]{cao2024survey}
    Cao, H.; Tan, C.; Gao, Z.; Xu, Y.; Chen, G.; Heng, P.-A.; and Li, S.~Z. 2024.
    \newblock A survey on generative diffusion models.
    \newblock \emph{IEEE Transactions on Knowledge and Data Engineering}.
    
    \bibitem[{Chen et~al.(2025)Chen, Zhao, Xu, and Lan}]{chen2025jointtuner}
    Chen, F.; Zhao, S.; Xu, C.; and Lan, L. 2025.
    \newblock JointTuner: Appearance-Motion Adaptive Joint Training for Customized Video Generation.
    \newblock \emph{arXiv preprint arXiv:2503.23951}.
    
    \bibitem[{Chen et~al.(2023)Chen, Jiang, Liu, Huang, Fu, Chen, and Yu}]{chen2023executing}
    Chen, X.; Jiang, B.; Liu, W.; Huang, Z.; Fu, B.; Chen, T.; and Yu, G. 2023.
    \newblock Executing your Commands via Motion Diffusion in Latent Space.
    \newblock In \emph{Proceedings of the IEEE/CVF Conference on Computer Vision and Pattern Recognition}, 18000--18010.
    
    \bibitem[{Chung et~al.(2021)Chung, Wuu, Yang, Tai, and Tang}]{chung2021haa500}
    Chung, J.; Wuu, C.-h.; Yang, H.-r.; Tai, Y.-W.; and Tang, C.-K. 2021.
    \newblock Haa500: Human-centric atomic action dataset with curated videos.
    \newblock In \emph{Proceedings of the IEEE/CVF international conference on computer vision}, 13465--13474.
    
    \bibitem[{Cideron et~al.(2024)Cideron, Girgin, Verzetti, Vincent, Kastelic, Borsos, McWilliams, Ungureanu, Bachem, Pietquin et~al.}]{cideron2024musicrl}
    Cideron, G.; Girgin, S.; Verzetti, M.; Vincent, D.; Kastelic, M.; Borsos, Z.; McWilliams, B.; Ungureanu, V.; Bachem, O.; Pietquin, O.; et~al. 2024.
    \newblock Musicrl: Aligning music generation to human preferences.
    \newblock \emph{arXiv preprint arXiv:2402.04229}.
    
    \bibitem[{Collins et~al.(2024)Collins, Kim, Bitton, Rieser, Omidshafiei, Hu, Chen, Dutta, Chang, Lee et~al.}]{collins2024beyond}
    Collins, K.~M.; Kim, N.; Bitton, Y.; Rieser, V.; Omidshafiei, S.; Hu, Y.; Chen, S.; Dutta, S.; Chang, M.; Lee, K.; et~al. 2024.
    \newblock Beyond Thumbs Up/Down: Untangling Challenges of Fine-Grained Feedback for Text-to-Image Generation.
    \newblock In \emph{Proceedings of the AAAI/ACM Conference on AI, Ethics, and Society}, volume~7, 293--303.
    
    \bibitem[{Dai et~al.(2024)Dai, Chen, Wang, Liu, Dai, and Tang}]{dai2024motionlcm}
    Dai, W.; Chen, L.-H.; Wang, J.; Liu, J.; Dai, B.; and Tang, Y. 2024.
    \newblock Motionlcm: Real-time controllable motion generation via latent consistency model.
    \newblock In \emph{European Conference on Computer Vision}, 390--408. Springer.
    
    \bibitem[{DeepMind(2025)}]{Gemini2Flash}
    DeepMind, G. 2025.
    \newblock Gemini 2.0 Flash.
    \newblock url = {https://deepmind.google/technologies/gemini/flash/}.
    \newblock Gemini 2.0 Flash model released by Google DeepMind.
    
    \bibitem[{Dynkin and Dynkin(1965)}]{dynkin1965markov}
    Dynkin, E.~B.; and Dynkin, E.~B. 1965.
    \newblock \emph{Markov processes}.
    \newblock Springer.
    
    \bibitem[{Ethayarajh et~al.(2024)Ethayarajh, Xu, Muennighoff, Jurafsky, and Kiela}]{ethayarajh2024kto}
    Ethayarajh, K.; Xu, W.; Muennighoff, N.; Jurafsky, D.; and Kiela, D. 2024.
    \newblock Kto: Model alignment as prospect theoretic optimization.
    \newblock \emph{arXiv preprint arXiv:2402.01306}.
    
    \bibitem[{Feng et~al.(2023)Feng, Liu, Yu, Yao, Hui, Guo, Lin, Xue, Shi, Li et~al.}]{feng2023dreamoving}
    Feng, M.; Liu, J.; Yu, K.; Yao, Y.; Hui, Z.; Guo, X.; Lin, X.; Xue, H.; Shi, C.; Li, X.; et~al. 2023.
    \newblock Dreamoving: A human video generation framework based on diffusion models.
    \newblock \emph{arXiv preprint arXiv:2312.05107}.
    
    \bibitem[{GPT4o(2024)}]{GPT4o}
    GPT4o. 2024.
    \newblock https://openai.com/index/hello-gpt-4o/.
    
    \bibitem[{Guo et~al.(2024)Guo, Mu, Javed, Wang, and Cheng}]{guo2024momask}
    Guo, C.; Mu, Y.; Javed, M.~G.; Wang, S.; and Cheng, L. 2024.
    \newblock Momask: Generative masked modeling of 3d human motions.
    \newblock In \emph{Proceedings of the IEEE/CVF Conference on Computer Vision and Pattern Recognition}, 1900--1910.
    
    \bibitem[{Guo et~al.(2022)Guo, Zou, Zuo, Wang, Ji, Li, and Cheng}]{guo2022generating}
    Guo, C.; Zou, S.; Zuo, X.; Wang, S.; Ji, W.; Li, X.; and Cheng, L. 2022.
    \newblock Generating diverse and natural 3d human motions from text.
    \newblock In \emph{Proceedings of the IEEE/CVF conference on computer vision and pattern recognition}, 5152--5161.
    
    \bibitem[{Ho, Jain, and Abbeel(2020)}]{ho2020denoising}
    Ho, J.; Jain, A.; and Abbeel, P. 2020.
    \newblock Denoising diffusion probabilistic models.
    \newblock \emph{Advances in neural information processing systems}, 33: 6840--6851.
    
    \bibitem[{Hu(2024)}]{hu2024animate}
    Hu, L. 2024.
    \newblock Animate anyone: Consistent and controllable image-to-video synthesis for character animation.
    \newblock In \emph{Proceedings of the IEEE/CVF Conference on Computer Vision and Pattern Recognition}, 8153--8163.
    
    \bibitem[{Jiang et~al.(2023{\natexlab{a}})Jiang, Chen, Liu, Yu, Yu, and Chen}]{jiang2023motiongpt}
    Jiang, B.; Chen, X.; Liu, W.; Yu, J.; Yu, G.; and Chen, T. 2023{\natexlab{a}}.
    \newblock Motiongpt: Human motion as a foreign language.
    \newblock \emph{Advances in Neural Information Processing Systems}, 36: 20067--20079.
    
    \bibitem[{Jiang et~al.(2023{\natexlab{b}})Jiang, Lu, Zhang, Ma, Han, Lyu, Li, and Chen}]{jiang2023rtmpose}
    Jiang, T.; Lu, P.; Zhang, L.; Ma, N.; Han, R.; Lyu, C.; Li, Y.; and Chen, K. 2023{\natexlab{b}}.
    \newblock Rtmpose: Real-time multi-person pose estimation based on mmpose.
    \newblock \emph{arXiv preprint arXiv:2303.07399}.
    
    \bibitem[{Khirodkar et~al.(2024)Khirodkar, Bagautdinov, Martinez, Zhaoen, James, Selednik, Anderson, and Saito}]{khirodkar2024sapiens}
    Khirodkar, R.; Bagautdinov, T.; Martinez, J.; Zhaoen, S.; James, A.; Selednik, P.; Anderson, S.; and Saito, S. 2024.
    \newblock Sapiens: Foundation for human vision models.
    \newblock In \emph{European Conference on Computer Vision}, 206--228. Springer.
    
    \bibitem[{Kong et~al.(2024)Kong, Tian, Zhang, Min, Dai, Zhou, Xiong, Li, Wu, Zhang et~al.}]{kong2024hunyuanvideo}
    Kong, W.; Tian, Q.; Zhang, Z.; Min, R.; Dai, Z.; Zhou, J.; Xiong, J.; Li, X.; Wu, B.; Zhang, J.; et~al. 2024.
    \newblock Hunyuanvideo: A systematic framework for large video generative models.
    \newblock \emph{arXiv preprint arXiv:2412.03603}.
    
    \bibitem[{Li et~al.(2025)Li, Xing, Sun, Ha, Shen, and Ho}]{li2025tokenmotion}
    Li, R.; Xing, D.; Sun, H.; Ha, Y.; Shen, J.; and Ho, C. 2025.
    \newblock TokenMotion: Decoupled Motion Control via Token Disentanglement for Human-centric Video Generation.
    \newblock In \emph{Proceedings of the Computer Vision and Pattern Recognition Conference}, 1951--1961.
    
    \bibitem[{Li et~al.(2024)Li, Yuan, He, Qiu, Zhu, Gu, Shen, Dong, Dong, and Yang}]{li2024lamp}
    Li, Z.; Yuan, W.; He, Y.; Qiu, L.; Zhu, S.; Gu, X.; Shen, W.; Dong, Y.; Dong, Z.; and Yang, L.~T. 2024.
    \newblock LaMP: Language-Motion Pretraining for Motion Generation, Retrieval, and Captioning.
    \newblock \emph{arXiv preprint arXiv:2410.07093}.
    
    \bibitem[{Liang et~al.(2024)Liang, Zhang, Li, Yu, and Xu}]{liang2024intergen}
    Liang, H.; Zhang, W.; Li, W.; Yu, J.; and Xu, L. 2024.
    \newblock Intergen: Diffusion-based multi-human motion generation under complex interactions.
    \newblock \emph{International Journal of Computer Vision}, 1--21.
    
    \bibitem[{Ma et~al.(2024)Ma, He, Cun, Wang, Chen, Li, and Chen}]{ma2024follow}
    Ma, Y.; He, Y.; Cun, X.; Wang, X.; Chen, S.; Li, X.; and Chen, Q. 2024.
    \newblock Follow your pose: Pose-guided text-to-video generation using pose-free videos.
    \newblock In \emph{Proceedings of the AAAI Conference on Artificial Intelligence}, volume~38, 4117--4125.
    
    \bibitem[{Peng et~al.(2024)Peng, Chen, Wang, Lu, and Qiao}]{peng2024conditionvideo}
    Peng, B.; Chen, X.; Wang, Y.; Lu, C.; and Qiao, Y. 2024.
    \newblock ConditionVideo: training-free condition-guided video generation.
    \newblock In \emph{Proceedings of the AAAI Conference on Artificial Intelligence}, volume~38, 4459--4467.
    
    \bibitem[{Raffel et~al.(2020)Raffel, Shazeer, Roberts, Lee, Narang, Matena, Zhou, Li, and Liu}]{raffel2020exploring}
    Raffel, C.; Shazeer, N.; Roberts, A.; Lee, K.; Narang, S.; Matena, M.; Zhou, Y.; Li, W.; and Liu, P.~J. 2020.
    \newblock Exploring the limits of transfer learning with a unified text-to-text transformer.
    \newblock \emph{Journal of machine learning research}, 21(140): 1--67.
    
    \bibitem[{Ramesh et~al.(2022)Ramesh, Dhariwal, Nichol, Chu, and Chen}]{ramesh2022hierarchical}
    Ramesh, A.; Dhariwal, P.; Nichol, A.; Chu, C.; and Chen, M. 2022.
    \newblock Hierarchical text-conditional image generation with clip latents.
    \newblock \emph{arXiv preprint arXiv:2204.06125}, 1(2): 3.
    
    \bibitem[{Ruiz et~al.(2023)Ruiz, Li, Jampani, Pritch, Rubinstein, and Aberman}]{ruiz2023dreambooth}
    Ruiz, N.; Li, Y.; Jampani, V.; Pritch, Y.; Rubinstein, M.; and Aberman, K. 2023.
    \newblock Dreambooth: Fine Tuning Text-to-Image Diffusion Models for Subject-Driven Generation.
    \newblock In \emph{Proceedings of the IEEE/CVF Conference on Computer Vision and Pattern Recognition}, 22500--22510.
    
    \bibitem[{Shafir et~al.(2023)Shafir, Tevet, Kapon, and Bermano}]{shafir2023human}
    Shafir, Y.; Tevet, G.; Kapon, R.; and Bermano, A.~H. 2023.
    \newblock Human motion diffusion as a generative prior.
    \newblock \emph{arXiv preprint arXiv:2303.01418}.
    
    \bibitem[{Song, Meng, and Ermon(2020)}]{song2020denoising}
    Song, J.; Meng, C.; and Ermon, S. 2020.
    \newblock Denoising diffusion implicit models.
    \newblock \emph{arXiv preprint arXiv:2010.02502}.
    
    \bibitem[{Soomro, Zamir, and Shah(2012)}]{soomro2012ucf101}
    Soomro, K.; Zamir, A.~R.; and Shah, M. 2012.
    \newblock UCF101: A dataset of 101 human actions classes from videos in the wild.
    \newblock \emph{arXiv preprint arXiv:1212.0402}.
    
    \bibitem[{Sun et~al.(2024)Sun, Zheng, Huang, Ma, Huang, and Hu}]{sun2024lgtm}
    Sun, H.; Zheng, R.; Huang, H.; Ma, C.; Huang, H.; and Hu, R. 2024.
    \newblock LGTM: Local-to-Global Text-Driven Human Motion Diffusion Model.
    \newblock In \emph{ACM SIGGRAPH 2024 Conference Papers}, 1--9.
    
    \bibitem[{Tan et~al.(2024)Tan, Gong, Wang, Zhang, Zheng, Zheng, Zheng, Chen, and Yang}]{tan2024AnimateX}
    Tan, S.; Gong, B.; Wang, X.; Zhang, S.; Zheng, D.; Zheng, R.; Zheng, K.; Chen, J.; and Yang, M. 2024.
    \newblock Animate-X: Universal Character Image Animation with Enhanced Motion Representation.
    \newblock \emph{arXiv preprint arXiv:2410.10306}.
    
    \bibitem[{Tevet et~al.(2023)Tevet, Raab, Gordon, Shafir, Cohen-or, and Bermano}]{tevet2023human}
    Tevet, G.; Raab, S.; Gordon, B.; Shafir, Y.; Cohen-or, D.; and Bermano, A.~H. 2023.
    \newblock Human Motion Diffusion Model.
    \newblock In \emph{The Eleventh International Conference on Learning Representations}.
    
    \bibitem[{Wallace et~al.(2024)Wallace, Dang, Rafailov, Zhou, Lou, Purushwalkam, Ermon, Xiong, Joty, and Naik}]{wallace2024diffusion}
    Wallace, B.; Dang, M.; Rafailov, R.; Zhou, L.; Lou, A.; Purushwalkam, S.; Ermon, S.; Xiong, C.; Joty, S.; and Naik, N. 2024.
    \newblock Diffusion model alignment using direct preference optimization.
    \newblock In \emph{Proceedings of the IEEE/CVF Conference on Computer Vision and Pattern Recognition}, 8228--8238.
    
    \bibitem[{Wan et~al.(2025)Wan, Wang, Ai, Wen, Mao, Xie, Chen, Yu, Zhao, Yang et~al.}]{wan2025wan}
    Wan, T.; Wang, A.; Ai, B.; Wen, B.; Mao, C.; Xie, C.-W.; Chen, D.; Yu, F.; Zhao, H.; Yang, J.; et~al. 2025.
    \newblock Wan: Open and advanced large-scale video generative models.
    \newblock \emph{arXiv preprint arXiv:2503.20314}.
    
    \bibitem[{Wang et~al.(2025)Wang, Wang, Ni, Zhao, Yang, Zhu, Zhang, Zhou, Chen, Huang et~al.}]{wang2025humandreamer}
    Wang, B.; Wang, X.; Ni, C.; Zhao, G.; Yang, Z.; Zhu, Z.; Zhang, M.; Zhou, Y.; Chen, X.; Huang, G.; et~al. 2025.
    \newblock HumanDreamer: Generating Controllable Human-Motion Videos via Decoupled Generation.
    \newblock \emph{arXiv preprint arXiv:2503.24026}.
    
    \bibitem[{Wang et~al.(2024{\natexlab{a}})Wang, Wang, Guo, Chen, Zhang, Ma, and Zheng}]{wang2024rlcoder}
    Wang, Y.; Wang, Y.; Guo, D.; Chen, J.; Zhang, R.; Ma, Y.; and Zheng, Z. 2024{\natexlab{a}}.
    \newblock Rlcoder: Reinforcement learning for repository-level code completion.
    \newblock \emph{arXiv preprint arXiv:2407.19487}.
    
    \bibitem[{Wang et~al.(2024{\natexlab{b}})Wang, Wang, Gong, Huang, He, Ouyang, Jiao, Feng, Dou, Tang et~al.}]{wang2024holistic}
    Wang, Y.; Wang, Z.; Gong, J.; Huang, D.; He, T.; Ouyang, W.; Jiao, J.; Feng, X.; Dou, Q.; Tang, S.; et~al. 2024{\natexlab{b}}.
    \newblock Holistic-motion2d: Scalable whole-body human motion generation in 2d space.
    \newblock \emph{arXiv preprint arXiv:2406.11253}.
    
    \bibitem[{Wang et~al.(2024{\natexlab{c}})Wang, Li, Zhu, Guo, Dou, and Li}]{wang2024customvideo}
    Wang, Z.; Li, A.; Zhu, L.; Guo, Y.; Dou, Q.; and Li, Z. 2024{\natexlab{c}}.
    \newblock Customvideo: Customizing text-to-video generation with multiple subjects.
    \newblock \emph{arXiv preprint arXiv:2401.09962}.
    
    \bibitem[{Xu et~al.(2024)Xu, Liu, Wu, Tong, Li, Ding, Tang, and Dong}]{xu2024imagereward}
    Xu, J.; Liu, X.; Wu, Y.; Tong, Y.; Li, Q.; Ding, M.; Tang, J.; and Dong, Y. 2024.
    \newblock Imagereward: Learning and Evaluating Human Preferences for Text-to-Image Generation.
    \newblock \emph{Advances in Neural Information Processing Systems}, 36.
    
    \bibitem[{Ye et~al.(2024)Ye, Xu, Liu, Hu, Yan, Qian, Zhang, Huang, and Zhou}]{ye2024mplug}
    Ye, J.; Xu, H.; Liu, H.; Hu, A.; Yan, M.; Qian, Q.; Zhang, J.; Huang, F.; and Zhou, J. 2024.
    \newblock mplug-owl3: Towards long image-sequence understanding in multi-modal large language models.
    \newblock \emph{arXiv preprint arXiv:2408.04840}.
    
    \bibitem[{Yu, Seo, and Son(2023)}]{yu2023zero}
    Yu, S.; Seo, P.~H.; and Son, J. 2023.
    \newblock Zero-shot referring image segmentation with global-local context features.
    \newblock In \emph{Proceedings of the IEEE/CVF conference on computer vision and pattern recognition}, 19456--19465.
    
    \bibitem[{Zhai et~al.(2024)Zhai, Lin, Li, Lin, Wang, Yang, Doermann, Yuan, Liu, and Wang}]{zhai2024idol}
    Zhai, Y.; Lin, K.; Li, L.; Lin, C.-C.; Wang, J.; Yang, Z.; Doermann, D.; Yuan, J.; Liu, Z.; and Wang, L. 2024.
    \newblock Idol: Unified dual-modal latent diffusion for human-centric joint video-depth generation.
    \newblock In \emph{European Conference on Computer Vision}, 134--152. Springer.
    
    \bibitem[{Zhang et~al.(2023)Zhang, Zhang, Zhang, and Kweon}]{zhang2023text}
    Zhang, C.; Zhang, C.; Zhang, M.; and Kweon, I.~S. 2023.
    \newblock Text-to-Image Diffusion Models in Generative AI: A Survey.
    \newblock \emph{arXiv preprint arXiv:2303.07909}.
    
    \bibitem[{Zhang, Rao, and Agrawala(2023)}]{zhang2023adding}
    Zhang, L.; Rao, A.; and Agrawala, M. 2023.
    \newblock Adding conditional control to text-to-image diffusion models.
    \newblock In \emph{Proceedings of the IEEE/CVF International Conference on Computer Vision}, 3836--3847.
    
    \bibitem[{Zhang et~al.(2024{\natexlab{a}})Zhang, Cai, Pan, Hong, Guo, Yang, and Liu}]{zhang2024motiondiffuse}
    Zhang, M.; Cai, Z.; Pan, L.; Hong, F.; Guo, X.; Yang, L.; and Liu, Z. 2024{\natexlab{a}}.
    \newblock Motiondiffuse: Text-driven human motion generation with diffusion model.
    \newblock \emph{IEEE Transactions on Pattern Analysis and Machine Intelligence}.
    
    \bibitem[{Zhang, Zhu, and Derpanis(2013)}]{zhang2013actemes}
    Zhang, W.; Zhu, M.; and Derpanis, K.~G. 2013.
    \newblock From actemes to action: A strongly-supervised representation for detailed action understanding.
    \newblock In \emph{Proceedings of the IEEE international conference on computer vision}, 2248--2255.
    
    \bibitem[{Zhang et~al.(2024{\natexlab{b}})Zhang, Gu, Wang, Wang, Cheng, Zhu, and Zou}]{zhang2024mimicmotion}
    Zhang, Y.; Gu, J.; Wang, L.-W.; Wang, H.; Cheng, J.; Zhu, Y.; and Zou, F. 2024{\natexlab{b}}.
    \newblock Mimicmotion: High-quality human motion video generation with confidence-aware pose guidance.
    \newblock \emph{arXiv preprint arXiv:2406.19680}.
    
    \bibitem[{Zhang et~al.(2024{\natexlab{c}})Zhang, Huang, Liu, Tang, Lu, Chen, Bai, Chu, Yu, and Ouyang}]{zhang2024motiongpt}
    Zhang, Y.; Huang, D.; Liu, B.; Tang, S.; Lu, Y.; Chen, L.; Bai, L.; Chu, Q.; Yu, N.; and Ouyang, W. 2024{\natexlab{c}}.
    \newblock Motiongpt: Finetuned llms are general-purpose motion generators.
    \newblock In \emph{Proceedings of the AAAI Conference on Artificial Intelligence}, volume~38, 7368--7376.
    
    \bibitem[{Zhang et~al.(2024{\natexlab{d}})Zhang, Liu, Reid, Hartley, Zhuang, and Tang}]{zhang2024motion}
    Zhang, Z.; Liu, A.; Reid, I.; Hartley, R.; Zhuang, B.; and Tang, H. 2024{\natexlab{d}}.
    \newblock Motion mamba: Efficient and long sequence motion generation.
    \newblock In \emph{European Conference on Computer Vision}, 265--282. Springer.
    
    \bibitem[{Zhao et~al.(2024)Zhao, Chen, Chen, Bao, Hao, Yuan, and Wong}]{zhao2024uni}
    Zhao, S.; Chen, D.; Chen, Y.-C.; Bao, J.; Hao, S.; Yuan, L.; and Wong, K.-Y.~K. 2024.
    \newblock Uni-controlnet: All-in-one control to text-to-image diffusion models.
    \newblock \emph{Advances in Neural Information Processing Systems}, 36.
    
    \end{thebibliography}
\end{document}